\documentclass[10pt,twocolumn,letterpaper]{article}

\usepackage{cvpr}              % To produce the CAMERA-READY version
\renewcommand{\paragraph}[1]{\vspace{.5em}\noindent\textbf{#1.}\hspace{0.5em}}

\usepackage{amsmath, amsfonts, amssymb}
\usepackage{pifont}   % For \ding symbols
\usepackage{booktabs}
\usepackage{multirow}
\usepackage{wrapfig} % in preamble
\usepackage[normalem]{ulem}

\usepackage{graphicx}
\setkeys{Gin}{draft=false} % force images to render
\usepackage[table]{xcolor}
\definecolor{mred}{RGB}{160,0,0}
\definecolor{mgreen}{RGB}{0,100,0}
\definecolor{mblue}{RGB}{0,51,179}
\definecolor{cvpurple}{RGB}{128,128,255}
\definecolor{grey}{RGB}{128,128,128}

\usepackage{amsmath, amsfonts, bm}

\def\1{\bm{1}}                           % indicator / all-ones vector

\DeclareMathAlphabet{\mathsfit}{\encodingdefault}{\sfdefault}{m}{sl}
\SetMathAlphabet{\mathsfit}{bold}{\encodingdefault}{\sfdefault}{bx}{n}
\def\sR{{\mathbb{R}}}        % real numbers
\DeclareMathOperator*{\argmax}{arg\,max}

\newcommand{\mat}[1]{\mathbf{#1}}                   % Matrix or 2D tensor (e.g., \mat{X})
\newcommand{\matdn}[2]{\mathbf{#1}_{#2}}            % Matrix with subscript (e.g., \matdn{X}{b} → X_b)
\newcommand{\matup}[2]{\mathbf{#1}^{\rm #2}}        % Matrix with superscript (e.g., \matup{X}{t} → X^t)
\newcommand{\matud}[3]{\mathbf{#1}^{\rm #2}_{#3}}   % Matrix with both super- and subscript (e.g., \matud{X}{t}{b} → X^t_b)

\newcommand{\vect}[1]{\boldsymbol{#1}}              % Vector (e.g., \vec{p})
\newcommand{\vecdn}[2]{\boldsymbol{#1}_{#2}}        % Vector with subscript (e.g., \vecdn{p}{i} → p_i)
\newcommand{\vecup}[2]{\boldsymbol{#1}^{\rm #2}}    % Vector with superscript (e.g., \vecup{p}{c} → p^c)
\newcommand{\vecud}[3]{\boldsymbol{#1}^{\rm #2}_{#3}} % Vector with both super- and subscript (e.g., \vecupdn{p}{c}{t} → p^c_t)

\newcommand{\spc}[1]{\mathcal{#1}}                  % General space or set (e.g., \spc{X} → ℳx)
\newcommand{\spcup}[2]{\mathcal{#1}^{\rm #2}}       % Space with subscript (e.g., \spcdn{X}{b} → x_b)
\newcommand{\fundn}[2]{\mathcal{#1}_{\rm #2}}               % function with subscript
\def\gfsprdgm{GFS-PCS\xspace} 
\def\ifsprdgm{IFS-PCS\xspace} 
\def\ourmethod{SCOPE\xspace}

\def\supp{\textbf{\textit{Supp.}\xspace}}

\newcommand{\cmark}{\color{mgreen}\ding{51}}  % check mark
\newcommand{\xmark}{\color{mred}{\ding{55}}}  % cross mark
\definecolor{cvprblue}{rgb}{0.21,0.49,0.74}
\usepackage[pagebackref,breaklinks,colorlinks,allcolors=cvprblue]{hyperref}

\def\paperID{} % *** Enter the Paper ID here
\def\confName{CVPR}
\def\confYear{2026}

\title{\texttt{\ourmethod}: Scene-Contextualized Incremental Few-Shot 3D Segmentation}

\author{
Vishal Thengane$^{1,2}$ \qquad
Zhaochong An$^{3}$ \qquad
Tianjin Huang$^{4}$ \qquad
Son Lam Phung$^{2}$ \qquad \\
Abdesselam Bouzerdoum$^{2}$ \qquad
Lu Yin$^{1}$ \qquad
Na Zhao$^{5,*}$ \qquad
Xiatian Zhu$^{1,}$\thanks{Corresponding authors: Na Zhao and Xiatian Zhu} \\
$^{1}$University of Surrey, UK \qquad
$^{2}$University of Wollongong, Australia \\
$^{3}$University of Copenhagen, Denmark \qquad
$^{4}$University of Exeter, UK \\
$^{5}$Singapore University of Technology and Design, Singapore \\
}

\begin{document}
\maketitle
\begin{abstract}

Incremental Few-Shot (IFS) segmentation aims to learn new categories over time from only a few annotations. Although widely studied in 2D, it remains underexplored for 3D point clouds.
Existing methods suffer from catastrophic forgetting or fail to learn discriminative prototypes under sparse supervision, and often overlook a key cue: novel categories frequently appear as unlabelled background in base-training scenes.
We introduce \textbf{\ourmethod} (\textit{\textbf{S}cene-\textbf{CO}ntextualised \textbf{P}rototype \textbf{E}nrichment}), a plug-and-play background-guided prototype enrichment framework that integrates with any prototype-based 3D segmentation method.
After base training, a class-agnostic segmentation model extracts high-confidence pseudo-instances from background regions to build a prototype pool. When novel classes arrive with few labelled samples, relevant background prototypes are retrieved and fused with few-shot prototypes to form enriched representations without retraining the backbone or adding parameters.
Experiments on ScanNet and S3DIS show that \ourmethod achieves SOTA performance, improving novel-class IoU by up to 6.98\% and 3.61\%, and mean IoU by 2.25\% and 1.70\%, respectively, while maintaining low forgetting. 
Code is available \href{https://github.com/Surrey-UP-Lab/SCOPE}{here}.

\end{abstract}
    
\section{Introduction}
\label{sec:intro}

\begin{figure}[t]
  \centering
  \includegraphics[width=\linewidth]{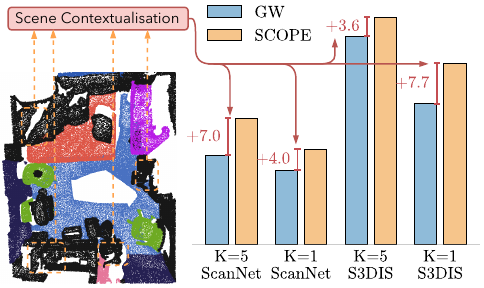}
  \caption{
    Compared with GW~\cite{xu2023generalized}, \ourmethod improves performance on novel classes by enriching prototypes with background context, without additional computational overhead.
  }
  \vspace{-0.45cm}
  \label{fig:teaser}
\end{figure}

Point Cloud Segmentation (PCS) forms the foundation of embodied perception tasks in indoor environments, 
% from service 
including robotics, autonomous driving, and AR/VR~\cite{Chen_2017_CVPR, engelcke2017vote3deep, milioto2019rangenet++, pierdicca2020point, Schult23ICRA}. 
% By assigning semantic labels to every point in a 3D scan, point cloud segmentation enables embodied systems to perceive and reason about their surroundings with fine-grained spatial understanding. 
While Fully Supervised (FS) methods achieve strong performance given abundant annotations~\cite{sun2022superpoint, choy20194d, zhao2021point}, practical deployments face two key constraints: (i)~\textit{novel categories emerge over time} as systems encounter new environments, and (ii)~\textit{only a handful of annotations} are available when such categories first appear.

% ===================Paragraph Start===================
% Recently, Few-Shot Class-Incremental (IFS) approaches~\cite{sur2025hyperbolic} have begun to bridge this gap, yet their performance under truly scarce supervision remains unsatisfactory.
% 
Existing approaches address these challenges independently.
Few-shot segmentation methods~\cite{zhao2021few, an2024multimodality} learn from limited examples but fail to retain previously learned knowledge. 
Generalised Few-Shot 3D PCS (\gfsprdgm) methods~\cite{tian2022generalized, xu2023generalized} recognise both base and novel classes, yet allow only a single update and assume prior knowledge of future categories--an unrealistic constraint in open-world scenarios.
In contrast, Class-Incremental 3D PCS (CI-PCS) approaches~\cite{li2017learning, kirkpatrick2017overcoming, thengane2025climb, Yang_2023_CVPR} support multiple updates but require abundant supervision and degrade sharply under limited annotations.
Incremental Few-Shot 3D PCS (\ifsprdgm) unifies these paradigms by enabling sequential learning of new classes from few-shot examples while preserving prior knowledge; however, despite progress in 2D~\cite{cermelli2021prototype, ganea2021incremental}, it remains underexplored for 3D point clouds. A comparison of these paradigms is presented in \cref{tab:method_comparision}.

% new classes on the fly from few examples without retrain- 068
% ing, which is particularly useful in exploration scenarios 069
% (e.g., robotics), where agents must incrementally recognise 070
% % newly encountered classes
% -------------------Paragraph End-------------------

% 
% ===================Paragraph Start===================
Directly applying these approaches proves suboptimal (\cref{tab:scannet-full}), as they struggle to generalise to novel classes under scarce supervision and often forget previously learned categories due to overfitting.
Conversely, GFS methods~\cite{tian2022generalized,xu2023generalized} retain base-class knowledge but show limited adaptability when learning new categories.
% strong 
% base-class retention but show limited adaptability 
% when learning
% to new categories.
%
% A recent hyperbolic-prototype approach for \ifsprdgm shows similar limitations, failing to achieve SOTA performance under truly low-shot conditions \cite{sur2025hyperbolic}.
Similarly, a recent hyperbolic-prototype approach for \ifsprdgm~\cite{sur2025hyperbolic} exhibits comparable limitations and does not yet match the strongest GFS baselines.
%
% We argue that the key 
% contributing 
We attribute these limitations to the underutilisation of background regions in base scenes, which often contain object-like structures indicative of future classes. 
% Effectively l
Leveraging this contextual information is therefore essential for robust \ifsprdgm.

\begin{table}[!tb]
\centering
% \small  % reduces font size to fit
\setlength{\tabcolsep}{3pt}  % adjusts column spacing
\caption{Comparison of PCS paradigms across future knowledge, few-shot, base-novel generalisation, and multi-step learning.}
\label{tab:method_comparision}
\resizebox{\linewidth}{!}{%
\begin{tabular}{lcccc}
  \toprule
    \textbf{PCS Paradigms} & \textbf{Future Knowledge} & \textbf{Few-Shot} & \textbf{Base + Novel} & \textbf{Multi-Step} \\
  \midrule
    Fully Supervised & -- & \xmark & -- & -- \\
    Few-Shot & \xmark & \cmark & \xmark & \xmark \\
    Generalized Few-Shot & \xmark & \cmark & \cmark & \xmark \\
    Class-Incremental & \cmark & \xmark & \cmark & \cmark \\
    \textbf{Incremental Few-Shot} & \cmark & \cmark & \cmark & \cmark \\
  \bottomrule
  \end{tabular}
}
\vspace{-0.4cm}
\end{table}
% -------------------Paragraph End-------------------

% 
% ===================Paragraph Start===================
To address this, we propose \textbf{\ourmethod} (\textit{\textbf{S}cene-\textbf{CO}ntextualised \textbf{P}rototype \textbf{E}nrichment}), a background-guided prototype enrichment framework.
Our key observation is that background regions--typically collapsed into a single coarse label--contain object-like structures that the backbone cannot disentangle during base training but often correspond to future classes.
In the \ifsprdgm setting, future class identities are unknown during the base stage; thus class-specific prototypes cannot be constructed in advance. We therefore employ an off-the-shelf class-agnostic segmentation model to extract high-confidence pseudo-instances and store them in an \textit{Instance Prototype Bank} (\textbf{IPB}), forming a reservoir of transferable object-level cues.
For each novel class, aligned context is retrieved from the IPB via the \textit{Contextual Prototype Retrieval} (\textbf{CPR}) module. Since retrieved instances vary in relevance, an \textit{Attention-Based Prototype Enrichment} (\textbf{APE}) mechanism selectively weights and fuses them with few-shot prototypes.
This produces context-aware and discriminative prototypes without modifying the backbone or introducing additional parameters, satisfying the minimal-adaptation principle of few-shot learning.
% -------------------Paragraph End-------------------

Compared with 
% state-of-the-art 
% SOTA
methods from related paradigms~\cite{tian2022generalized, sur2025hyperbolic, thengane2025climb}, 
% and built upon GW~\cite{xu2023generalized}, 
\ourmethod achieves consistent gains across benchmarks, demonstrating stronger novel-class adaptation and improved retention of prior knowledge.
In summary, our contributions are threefold:
(1) we propose a plug-and-play framework that mines contextual cues from base scenes and constructs an IPB using an off-the-shelf class-agnostic segmentation model with no additional computational overhead;
(2) we introduce a CPR module to retrieve relevant background cues without future class knowledge and an APE mechanism for prototype enrichment; and
(3) we establish new SOTA performance across multiple \ifsprdgm settings on standard 3D segmentation benchmarks.
\section{Related Work}
\label{sec:related}

\subsection{3D Scene Understanding}
Fully supervised 3D semantic segmentation has been extensively studied, typically requiring dense point-wise labels~\cite{hu2020randla, qi2017pointnet, qi2017pointnet++, wang2019dynamic, li2018pointcnn, zhao2021point}.  
Early models such as PointNet~\cite{qi2017pointnet} processed raw point clouds but struggled to capture local structures, while subsequent architectures including PointNet++~\cite{qi2017pointnet++}, DGCNN~\cite{wang2019dynamic}, and transformer-based models~\cite{zhao2021point} progressively enhanced geometric and contextual reasoning.  
Recently, foundation-style models~\cite{chen2023clip2scene, huang2023clip2point, jatavallabhula2023conceptfusion, wang2022ofa, zhang2023clip, zhou2024point} have explored cross-modal representations and open-vocabulary objectness without explicit class labels~\cite{huang2024segment3d, yang2024regionplc, takmaz2023openmask3d, peng2023openscene}.  
However, these approaches rely on large-scale annotations and fixed label spaces, limiting scalability in open-world environments.

\subsection{Few-Shot 3D Segmentation}
Few-shot 3D segmentation reduces reliance on large-scale annotated data. \citet{zhao2021few} proposed the first few-shot 3D segmentation method, enabling recognition of unseen classes from limited labelled samples. Existing approaches primarily refine prototypes or query embeddings via non-parametric optimisation~\cite{he2023prototype, mao2022bidirectional, xu2023towards, wang2023few, zhang2023few, zhu2024no, wei2024gandpfewshot}, or explicitly model support-query correlations~\cite{an2024rethinking, an2024multimodality}, as demonstrated by COSeg~\cite{an2024rethinking}.
Generalised few-shot methods jointly recognise base and novel classes. Representative examples include PIFS~\cite{cermelli2021prototype}, which refines prototypes via distillation; CAPL~\cite{tian2022generalized}, which incorporates co-occurrence priors; and GW~\cite{xu2023generalized}, which extends the paradigm to 3D via geometric cues.
Tsai \etal~\cite{tsaipseudo} mine background regions into pseudo-class prototypes but rely on language guidance for pre-clustering. In contrast, our method learns future classes across multiple stages without additional supervision, storing reusable background knowledge in a prototype pool.

\subsection{Incremental 3D Segmentation}
Incremental learning acquires new classes sequentially while mitigating catastrophic forgetting.
This is typically achieved through replay~\cite{rebuffi2017icarl}, knowledge distillation~\cite{hinton2015distilling}, regularisation~\cite{li2017learning}, or consolidation strategies~\cite{kirkpatrick2017overcoming, serra2018overcoming, aljundi2018memory, buzzega2020dark, cha2021co2l}.  
\citet{su2024balanced} and subsequent works~\cite{kontogianni2024continual, boudjoghra20243d} adapt these strategies to PCS, demonstrating promising results but often relying on large memory buffers and repeated retraining cycles.  
Thengane \etal~\cite{thengane2025climb} address long-tail classes via object-count priors, yet their approach still depends on abundant supervision for stable adaptation.  
As these methods are primarily designed for large-scale data regimes, they remain suboptimal under few-shot conditions.

\subsection{Incremental Few-Shot 3D Segmentation}
Incremental Few-Shot (IFS) learning
% , also known as FSCIL, 
aims to learn new classes sequentially from limited supervision while retaining prior knowledge. Although extensively studied in 2D vision~\cite{cermelli2021prototype, ganea2021incremental, liu2023learning}, its extension to 3D remains underexplored.
% , particularly for dense point-level prediction.  
In 3D, \citet{sur2025hyperbolic} introduce hyperbolic prototypes for IFS segmentation but remain behind \gfsprdgm baselines in performance.  
Our method addresses this gap by leveraging background context as transferable knowledge, achieving stronger performance for novel classes in the \ifsprdgm setting than methods from other paradigms.
\section{Methodology}

\begin{figure*}[!phtb]
  \centering
  \includegraphics[width=\linewidth]{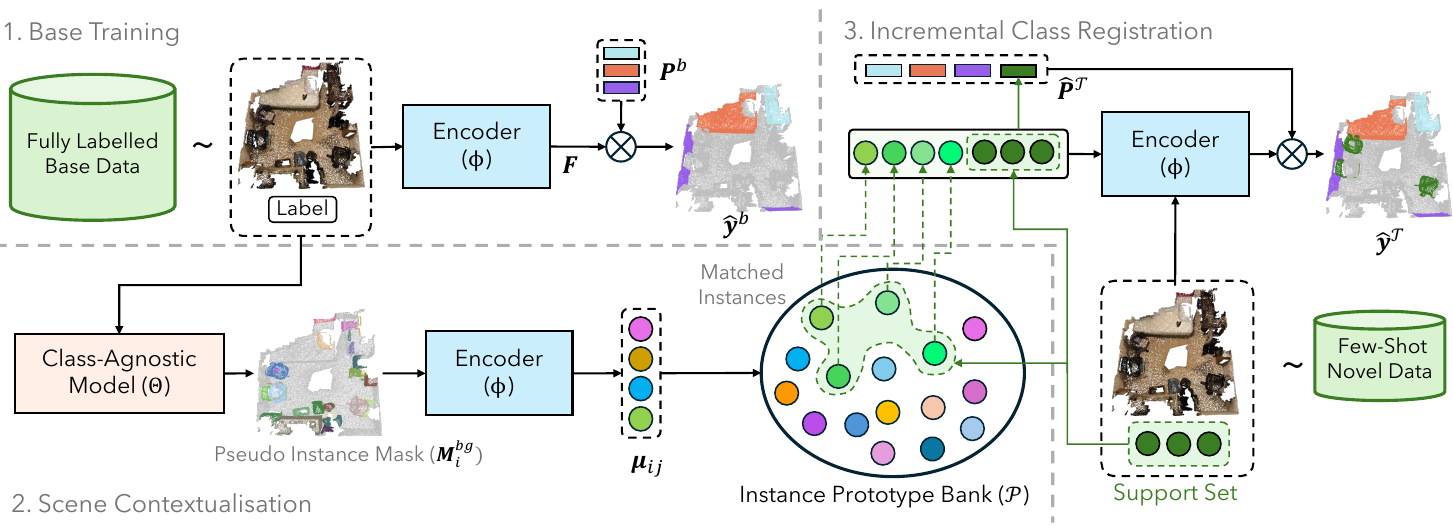}
    \caption{
        \textbf{Overview of \ourmethod}. 
        The framework comprises three stages: 
        (1) \textit{Base Training}, where the encoder~$\Phi$ is trained on labelled base data; 
        (2) \textit{Scene Contextualisation}, which extracts background regions to build an Instance Prototype Bank (IPB); and 
        (3) \textit{Incremental Class Registration}, where retrieved prototypes are fused with few-shot prototypes via attention to yield refined novel-class representations.
    }
    \vspace{-0.4cm}
  \label{fig:method}
\end{figure*}

\subsection{Problem Formulation}

The goal of the \ifsprdgm is to learn a model $\Phi$ that assigns each point in a point cloud to its % corresponding 
semantic label in a set $\vect{C}$, where $\vect{C}$ is progressively expanded using a small number of annotated samples for newly encountered classes.

\paragraph{\gfsprdgm}
% 
% Following the \gfsprdgm~\cite{xu2023generalized, an2025generalized} paradigm, l
Let $\vecup{C}{b}$ and $\vecup{C}{n}$ denote disjoint sets of base and novel classes, respectively, such that $\vect{C} = \vecup{C}{b} \cup \vecup{C}{n}$ and $|\vecup{C}{b}| = N^{b}$, $|\vecup{C}{n}| = N^{n}$, giving $N = N^{b} + N^{n}$ classes.
The model is first trained on the base classes using  
$
% \matup{D}{b} = 
%     \{(\mat{X},\, \vecup{y}{b}) 
%     \mid \mat{X} \in \spc{X},\, \vecup{y}{b} \in \spcup{Y}{b} \subseteq \vecup{C}{b}\}
\matup{D}{b} = 
    \{(\mat{X},\, \vecup{y}{b}) 
    \mid \mat{X} \in \spc{X},\, \vecup{y}{b} \in \spcup{Y}{b}\},
$
where $\vecup{y}{b}$ assigns each point a label from $\vecup{C}{b}$ or background ($-1$).
Novel classes are introduced in a single batch, each with $K$ labelled support examples, forming the few-shot dataset
$
\matup{D}{n} = 
\big\{\{ (\mat{X},\, \vecud{y}{c}{k}) \}_{k=1}^{K} \big\}_{c \in \vecup{C}{n}}
$. 
Each support point cloud 
$\matdn{X}{k} \in \mathbb{R}^{M \times d_0}$ contains $M$ points with $d_0$-dimensional features, and 
$\vecud{y}{c}{k}$ provides point-wise annotations for class~$c$ with all other points labelled as background.
Evaluation is performed on $\matup{D}{\mathrm{test}}$ across all classes in $\vect{C}$.

\paragraph{\ifsprdgm}  
% Different from \gfsprdgm, in the 
In contrast, in the \ifsprdgm setting, novel classes are registered over multiple stages, reflecting a naturally evolving environment.
Let $\vecup{C}{t}$ denote the set of semantic classes known after stage $t$, where each stage expands the model's output space.  
During learning, the model sequentially receives datasets $\{\matup{D}{0}, \ldots, \matup{D}{T}\}$, where  
$\matup{D}{t} = \{(\mat{X},\, \vecup{y}{t}) \mid \mat{X} \in \spc{X},\, \vecup{y}{t} \in \spcup{Y}{t}\}$.  
In each $\matup{D}{t}$, $\mat{X} \in \sR^{M \times d_0}$ denotes a point cloud with $M$ points and $d_0$-dimensional features, while $\vecup{y}{t} \in \sR^{M}$ assigns one class index per point, $\spcup{Y}{t} \subseteq \vecup{C}{t}$.  
The dataset $\matup{D}{0}$ contains abundant base-class examples, analogous to $\matup{D}{b}$ in \gfsprdgm.
Subsequent stages introduce novel classes incrementally via few-shot support datasets,
$\matup{D}{n} = \bigcup_{t = 1}^{T} \matup{D}{t}$,
each extending the known class set.
Although the same point cloud $\mat{X}$ may reappear across stages, only labels corresponding to the currently known classes $\vecup{C}{t}$ are available, \ie, supervision is restricted to classes introduced at that stage\footnote{Henceforth, we use $t{=}0$ and $t{=}b$ interchangeably to denote the $0$-th stage of the \ifsprdgm\ setting.}.

At $t = 0$, the model is trained on the large base dataset $\matup{D}{0} = \{(\matdn{X}{i}, \vecud{y}{0}{i})\}_{i=1}^{|\matup{D}{0}|}$ with labels drawn from $\vecup{C}{0}$.  
For each subsequent stage $t \geq 1$, the model receives a small few-shot support set $\matup{D}{t}$ introducing previously unseen classes from $\vecup{C}{t}$.  
Specifically, each novel class $c \in \vecup{C}{t}$ is supported by exactly $K$ labelled point clouds:
\begin{equation}
\big\{(\matdn{X}{k},\, \vecud{y}{c}{k})\big\}_{k=1}^{K}, \quad 
\matdn{X}{k} \in \sR^{M \times d_0},\ \vecud{y}{c}{k} \in \sR^{M}.
\label{eq:nway-kshot}
\end{equation}
Each $\vecud{y}{c}{k}$ assigns label $c$ to points of that class and background to all others, \ie,  
$\vecud{y}{c}{k}(x) \in \{c, -1\}, \forall\, x \in \matdn{X}{k},$  
where $-1$ denotes background or unknown regions.  
Evaluation is conducted on a test set $\matup{D}{\text{test}}$ covering all categories observed so far, \ie, 
$\matup{Y}{\text{test}} \subseteq \vecup{C}{t'}$, 
where $\vecup{C}{t'} = \bigcup_{i=0}^{t} \vecup{C}{i}$.

\subsection{Method Overview}

\paragraph{Our Idea}
The core idea of \ourmethod is to exploit contextual cues in base scenes to improve incremental few-shot point cloud segmentation.  
Although base training focuses only on known categories, background regions contain rich geometric and semantic signals corresponding to objects that later emerge as novel classes. 
However, directly extracting background features with the encoder $\Phi$ yields coarse, non-discriminative embeddings, as all unseen regions are collapsed into a single background class.
To address this, we incorporate a class-agnostic segmentation model $\Theta$~\cite{huang2024segment3d} to detect object-like regions in the background, mining latent structure to produce transferable contextual cues that enrich few-shot representations in later stages.
This allows adaptation to new classes without introducing additional parameters or retraining the encoder, effectively bridging few-shot generalisation and incremental learning in PCS.

% ------------------------------------------------------------
% Method Overview
% ------------------------------------------------------------

\paragraph{Overview}
\ourmethod is implemented as a three-stage pipeline that operationalises this idea through background mining and prototype refinement.
As illustrated in \cref{fig:method}, the framework consists of: (i) Base Training, (ii) Scene Contextualisation, and (iii) Incremental Class Registration.
During \textit{Base Training}, the encoder~$\Phi$ is learned together with base prototypes $\matup{P}{b}$ on fully labelled base data.
The \textit{Scene Contextualisation} stage applies the class-agnostic model~$\Theta$ to background regions and converts the resulting masks into an \textit{Instance Prototype Bank} (IPB), denoted $\spc{P}$.
During \textit{Incremental Class Registration}, each novel class $c \in \vecup{C}{t}$ is initialised with a few-shot prototype $\vecup{p}{c}$.
Our \textit{Contextual Prototype Retrieval} (CPR) module retrieves a class-specific subset $\spcup{B}{c} \subset \spc{P}$ of semantically aligned prototypes, which are fused via the \textit{Attention-based Prototype Enrichment} (APE) module to obtain enriched representations $\vecup{\tilde{p}}{c}$.
This enables expansion to new categories without modifying the encoder or introducing additional learnable parameters.
The framework remains fully plug-and-play and can be seamlessly applied to any prototype-based segmenter without altering the backbone or training schedule.

% =====================
% Base Training Stage
% =====================
\subsection{Base Training}
During the base stage, the model learns to encode geometric and semantic cues 
from the fully labelled base dataset $\matup{D}{b}$, building a discriminative embedding space.
It jointly trains a backbone network $\Phi^\prime$ and a projection head $\spc{H}$,
forming the overall encoder $\Phi = \spc{H} \circ \Phi^\prime$.
Given an input point cloud $\matdn{X}{i}$, the encoder produces point-wise embeddings:
\begin{equation}
    \matdn{F}{i} = \Phi(\matdn{X}{i}) = \spc{H}\big(\Phi^\prime(\matdn{X}{i})\big),
    \quad
    \matdn{F}{i} \in \mathbb{R}^{M \times D},
    \label{eq:embedding}
\end{equation}
where $M$ denotes the number of points in $\matdn{X}{i}$ and $D$ is the embedding dimension.
To associate these embeddings with semantic meaning, the model learns a set of base prototypes
$\matup{P}{b} = \{\vecup{p}{c}\}_{c \in \vecup{C}{b}} \in \mathbb{R}^{N^b \times D}$,
where each $\vecup{p}{c}$ acts as a class-level representative in the embedding space.
Per-point logits are obtained by measuring similarity between embeddings and prototypes:
% , and the similarity between point embeddings and prototypes yields per-point logits:
\begin{equation}
    \vecud{\Hat{y}}{b}{i} 
    = \argmax_{c \in \vecup{C}{b}}
    \big(\matdn{F}{i} \cdot (\matup{P}{b})^\top\big),
    \quad
    \vecud{\Hat{y}}{b}{i} \in \mathbb{R}^{M \times N^b}.
\end{equation}
%
% The model thus learns an embedding space where per-point features are aligned with their corresponding semantic prototypes.
Unlike prior methods that discard background features, our approach retains and reuses them to build transferable contextual prototypes for incremental adaptation.

% 
% The parameters of both $\Phi$ and $\spcup{H}{b}$ are optimised using a point-wise cross-entropy loss over the base classes $\vecup{C}{b}$:
% \begin{equation}
%     \mathcal{L}_{\mathrm{base}} = -\frac{1}{M} \sum_{p=1}^{M} \sum_{c \in \vecup{C}{b}} 
%     \mathbf{1}[y_p = c] \, \log \matup{Z}{b}_{p,c},
% \end{equation}
% where $\1_{[y_p = c]}$ is the indicator function and $\matup{Z}{b}_{p,c}$ denotes the predicted probability of point $p$ belonging to class $c$. 
% This training objective ensures that the backbone learns discriminative and semantically consistent features for the base classes, forming a strong foundation for subsequent incremental adaptation.
% % 

% =======================
% Scene Contextualisation
% =======================
\subsection{Scene Contextualisation}
After base training, the model predicts labels only from the base class set~$\vecup{C}{b}$, assigning all other points to the background.
However, these background points often belong to unlabelled objects or structural patterns that may correspond to future novel classes.
Because all unseen content is collapsed into a single background label, the encoder~$\Phi$ cannot delineate object boundaries, resulting in coarse and non-discriminative representations.
To address this limitation, we introduce the \textit{Scene Contextualisation Module} (\textbf{SCM}), which mines object-like regions using an off-the-shelf class-agnostic model~\cite{huang2024segment3d}. The model is applied once offline and subsequently discarded, and the resulting regions are converted into transferable prototypes.

\paragraph{Pseudo-Mask Generation} 
For each input scene $\matdn{X}{i}$ from the base dataset $\matup{D}{b}$, the class-agnostic segmentation model $\Theta$ predicts, in an offline manner, a set of pseudo-instance masks with associated confidence scores:
\begin{equation}
    \Theta(\matdn{X}{i}) 
    = \big\{\,(\matdn{\Hat{M}}{i,j},\, \vecdn{s}{i,j})\,\big\}_{j=1}^{Q_i},
\end{equation}
where $\matdn{\Hat{M}}{i,j} \in \{0,1\}^{M}$ denotes the $j$-th binary mask predicted for scene $\matdn{X}{i}$, $\vecdn{s}{i,j} \in [0,1]$ represents its corresponding confidence score, and $Q_i$ is the total number of predicted masks for that scene.
Only masks corresponding to background points and having confidence above a threshold $\tau$ are retained:
\begin{equation}
    \matud{M}{bg}{i} =
    \big\{\,
        \matdn{\Hat{M}}{i,j}
        \;\big|\;
        \matdn{\Hat{M}}{i,j} \subseteq \matdn{X}{i}[\vecud{y}{b}{i} = -1],
        \;
        \vecdn{s}{i,j} > \tau
    \,\big\}, 
\end{equation}
where $\matdn{X}{i}[\vecud{y}{b}{i} = -1]$ denotes the subset of points in scene $\matdn{X}{i}$ labelled as background in the base dataset.
Each resulting background mask $\matud{M}{bg}{i}$ delineates high-confidence object-like regions within the background, which are treated as potential instances of unseen or novel classes.

\paragraph{Instance Prototype Bank (IPB)}
The IPB aims to construct robust feature prototypes from the pseudo background masks 
$\matud{M}{bg}{i}$ predicted by the class-agnostic model~$\Theta$, providing transferable cues that facilitate subsequent novel-class registration.
Each input scene $\matdn{X}{i}$ is processed by the encoder~$\Phi$ to extract point-wise features 
$\matdn{F}{i}$ (\cref{eq:embedding}).
These features are then aggregated into instance-level prototypes for each pseudo mask 
$\matdn{\Hat{M}}{i,j} \in \matud{M}{bg}{i}$ using a masked average-pooling operator 
$\fundn{F}{Pool}$:
%
% \begin{equation}
% \begin{split}
%     \matdn{F}{i} &= \Phi(\matdn{X}{i}) \in \mathbb{R}^{M \times D}, \\
%     \vecdn{\mu}{i,j} &=
%     \frac{\sum_{p=1}^{M} \matdn{\Hat{M}}{i,j}[p] \, \matdn{F}{i}[p]}
%          {\sum_{p=1}^{M} \matdn{\Hat{M}}{i,j}[p]},
%     \quad
%     \matdn{\Hat{M}}{i,j} \in \{0,1\}^{M}.
% \end{split}
% \label{eq:prototype}
% \end{equation}
\begin{equation}
    \vecdn{\mu}{i,j} =
    \fundn{F}{Pool}\!\big(\matdn{F}{i},\, \matdn{\Hat{M}}{i,j}\big),
    \qquad
    \matdn{\Hat{M}}{i,j} \in \{0,1\}^{M}.
    \label{eq:prototype}
\end{equation}
The resulting vector $\vecdn{\mu}{i,j} \in \mathbb{R}^{D}$ serves as the \textit{instance prototype} for the
$j$-th pseudo background region of scene~$i$.  
% For simplicity, the operation in \cref{eq:prototype} is denoted as $\fundn{F}{Pool}$, such that $\vecdn{\mu}{i,j} = \fundn{F}{Pool}(\matdn{X}{i},\, \matud{M}{bg}{i})$.
Since the novel classes are unknown during base training, all such prototypes are collected across the full dataset to form the 
\textit{Instance Prototype Bank} (IPB):
\begin{equation}
    \spc{P} = \bigcup_{i} \bigcup_{j} \big\{\, \vecdn{\mu}{i,j} \,\big\}.
\end{equation}
For notational convenience, we collapse the scene-instance indices $(i,j)$ into a single bank index $b$, and denote the IPB elements as $\{\vecdn{\mu}{b}\}_{b=1}^{|\spc{P}|}$.
This bank provides a rich reservoir of object-like background patterns, enabling initialisation and refinement of novel-class representations during subsequent incremental stages.
Importantly, the IPB is constructed once after base training and then frozen throughout all incremental stages, introducing no additional optimisation or memory overhead during incremental adaptation.

% ==============================
% Incremental Class Registration
% ==============================
\subsection{Incremental Class Registration}
At each incremental stage~$t$ ($t \ge 1$), the few-shot dataset~$\matup{D}{t}$ introduces a set of 
novel categories~$\vecup{C}{t}$, each provided with $K$ annotated support point clouds 
$\{(\matdn{X}{k},\, \vecud{y}{c}{k})\}_{k=1}^{K}$.
For every novel class~$c \in \vecup{C}{t}$, an initial class prototype~$\vecup{p}{c}$ is obtained by 
extracting encoder features from the support examples and aggregating those corresponding to class~$c$:
\begin{equation}
    \vecup{p}{c} = 
    \frac{1}{K} \sum_{k=1}^{K} 
    \fundn{F}{Pool}\!\Big( \matdn{F}{k},\, \mathbf{1}\!\left[\vecdn{y}{k}{=}c\right] \Big),
    \label{eq:fpool-fewshot}
\end{equation}
%
% Here, $\fundn{F}{Pool}(\cdot,\cdot)$ denotes the masked average-pooling operator defined in~\cref{eq:prototype}, 
% which aggregates point-wise features belonging to class~$c$ within each support example.
%
where $\matdn{F}{k}$
% $ = \Phi(\matdn{X}{k})$ 
denotes the features of the $k$-th support example,
and $\mathbf{1}[\vecdn{y}{k}{=}c]$ is a binary mask selecting points labelled as class~$c$.
The resulting set of few-shot prototypes for stage~$t$ is denoted as
$\matup{P}{t} = \{\vecup{p}{c} \mid c \in \vecup{C}{t}\} \in \mathbb{R}^{|\vecup{C}{t}| \times D}$,
and provides the initial representation for each novel category before contextual refinement.

\paragraph{Contextual Prototype Retrieval (CPR)}
Few-shot prototypes $\vecup{p}{c}$, derived from limited annotations, often lack semantic diversity and do not fully capture the underlying class structure.
To compensate for this limitation, the \textit{Contextual Prototype Retrieval} (CPR) module identifies semantically aligned background embeddings from the IPB ($\spc{P}$).

For each novel class $c \in \vecup{C}{t}$, we compute the cosine
similarity between its few-shot prototype $\vecup{p}{c}$ and every background prototype $\vecdn{\mu}{b}$ in the IPB:
\begin{equation}
    \vecud{\sigma}{c}{b} =
    \frac{
        (\vecup{p}{c})^{\!\top}\vecdn{\mu}{b}
    }{
        \|\vecup{p}{c}\|_2 \,\|\vecdn{\mu}{b}\|_2
    }.
\end{equation}
The top-$R$ most relevant background prototypes are then selected to construct a class-specific context pool:
\begin{equation}
    \spcup{B}{c} =
    \big\{
        \vecdn{\mu}{b} \;\big|\;
        b \in \rm{TopR}(\vecud{\sigma}{c}{b})
    \big\}
    =
    \{\vecud{\mu}{c}{r}\}_{r=1}^{R}.
\end{equation}

This retrieval step yields a compact and semantically aligned set of background pseudo-instance prototypes that provide auxiliary structural cues for refining the initial few-shot prototypes for robust adaptation.
However, not all retrieved background prototypes are equally informative; some may be noisy or lack strong objectness.
Therefore, a dedicated enrichment mechanism is introduced to selectively integrate the most relevant contextual signals.

% 
% ==============================
% Attention-Based Prototype Enrichment
% ==============================

\paragraph{Attention-Based Prototype Enrichment (APE)}
During incremental stage~$t$, the retrieved contextual set for class~$c$ is 
$\spcup{B}{c} = \{\vecud{\mu}{c}{r}\}_{r=1}^{R}$.
Given the few-shot prototype $\vecup{p}{c}$ for $c \in \vecup{C}{t}$, we refine it through a parameter-free, attention-driven mechanism that fuses it with the these retrieved prototypes.

Both the few-shot and contextual prototypes are $\ell_{2}$-normalised:
\[
\vecup{\bar{p}}{c} = \fundn{F}{Norm}(\vecup{p}{c}),
\qquad
\spcup{\bar{B}}{c}
=
\big\{
    \fundn{F}{Norm}(\vecud{\mu}{c}{r})
    \,\big|\,
    \vecud{\mu}{c}{r} \in \spcup{B}{c}
\big\}.
\]
A scaled dot-product cross-attention operation then uses the few-shot prototype as the query and the contextual prototypes as keys and values, without introducing any learnable parameters or projection heads.
This produces a set of scalar attention weights, one per retrieved prototype, quantifying its relevance to class~$c$.
The context-enhanced representation is computed as:
\begin{equation}
    \vecup{h}{c} =
    \sum_{r=1}^{R}
    \mathrm{CrossAttention}\!\left(
        \vecup{\bar{p}}{c},\,
        \spcup{\bar{B}}{c}
    \right)_{r}
    \,
    \vecud{\bar{\mu}}{c}{r},
    \label{eq:proto-context}
\end{equation}

The enriched class prototype is obtained by combining the original few-shot prototype with the attention-weighted contextual prototypes:
\begin{equation}
    \vecup{\tilde{p}}{c} =
    \lambda\,\vecup{p}{c}
    + (1-\lambda)\,\vecup{h}{c},
    \qquad
    \lambda \in [0,1].
\end{equation}
The parameter-free attention mechanism suppresses noisy context while preserving transferable structural cues for prototype refinement.
The refined prototypes for all classes observed up to stage~$t$ are assembled as
\begin{equation}
    \matup{\tilde{P}}{\le t} =
    \big[\,\matup{P}{b},\ \ldots,\ \matup{\tilde{P}}{t}\,\big],
    \qquad
    \matup{\tilde{P}}{t} =
    \big\{
        \vecup{\tilde{p}}{c}
        \;\big|\;
        c \in \vecup{C}{t}
    \big\},
    \label{eq:refined-proto-set}
\end{equation}
where $\matup{P}{b}$ denotes the base prototypes and $\vecup{C}{t}$ the novel classes introduced at stage~$t$.
The cumulative class set up to stage~$t$ is $\vecup{C}{\le t} = \bigcup_{i=0}^{t}\vecup{C}{i}$, and the prototype matrix $\matup{\tilde{P}}{\le t}$ serves as the unified classifier for segmentation. The final point-wise predictions are obtained via:
\begin{equation}
    \vecud{\Hat{y}}{\le t}{i} =
    \argmax_{c \in \vecup{C}{\le t}}
    \big(
        \matdn{F}{i}
        \cdot
        (\matup{\tilde{P}}{\le t})^\top
    \big)
    \in \mathbb{R}^{M \times |\vecup{C}{\le t}|}.
\end{equation}
% {\color{mred}
% \input{_secs/3_method_backup}}
\section{Experiments}
We evaluate \ourmethod on two benchmarks against representative incremental, few-shot, and generalised few-shot baselines. We first present the experimental setup, followed by quantitative and qualitative results, ablations, and a discussion of efficiency and limitations (see \supp~for details).

\begin{table*}[!thb]
  \centering
    \caption{
        Comparison of baseline methods and \ourmethod on the \textit{ScanNet} dataset under the \ifsprdgm setting with $K{=}5$ and $K{=}1$.
        We report mIoU, mIoU-B, mIoU-N, HM, mIoU-I, and FPP. All metrics are higher is better except FPP. Best results are highlighted in \textbf{bold}.
    }
  \setlength{\tabcolsep}{3.5pt}
  \resizebox{\textwidth}{!}{
  \begin{tabular}{l r|rrrrrr|rrrrrr}
    \toprule
    \multirow{2}{*}{\textbf{Methods}} & \multirow{2}{*}{\textbf{Venue}} 
    & \multicolumn{6}{c|}{\textbf{$K{=}5$}} 
    & \multicolumn{6}{c}{\textbf{$K{=}1$}} \\
    \cmidrule(lr){3-8}\cmidrule(lr){9-14}
    & 
    & \textbf{mIoU} & \textbf{mIoU-B} & \textbf{mIoU-N} & \textbf{HM} & \textbf{mIOU-I} & \textbf{FPP}
    & \textbf{mIoU} & \textbf{mIoU-B} & \textbf{mIoU-N} & \textbf{HM} & \textbf{mIOU-I} & \textbf{FPP} \\
    \midrule
    \rowcolor{gray!15} JT (Oracle) & -- 
    & 45.34 & 48.68 & 36.97 & 42.03 & -- & -- 
    & 45.34 & 48.68 & 36.97 & 42.03 & -- & -- \\ 
    Finetuning & -- 
    & 0.38 & 0.48 & 0.15 & 0.23 & 12.28 & 34.25 
    & 0.42 & 0.49 & 0.26 & 0.34 & 10.93 & 40.67 \\
    LwF~\cite{li2017learning} & TPAMI'17 
    & 4.12 & 5.77 & 0.27 & 0.51 & 14.75 & 35.39 
    & 0.37 & 0.33 & 0.47 & 0.39 & 12.06 & 40.84 \\ 
    CLIMB-3D~\cite{thengane2025climb} & BMVC'25 
    & 4.23 & 5.92 & 0.02 & 0.04 & 15.98 & 35.25 
    & 0.02 & 0.00 & 0.06 & 0.00 & 10.36 & 41.16 \\
    EWC~\cite{kirkpatrick2017overcoming} & PNAS'17 
    & 3.87 & 5.41 & 0.03 & 0.06 & 17.66 & 35.76 
    & 2.33 & 4.42 & 0.02 & 0.05 & 16.57 & 35.22 \\ 
    GUA~\cite{Yang_2023_CVPR} & CVPR'23 
    & 13.43 & 18.01 & 3.14 & 5.35 & 24.59 & 25.55 
    & 8.97 & 11.69 & 1.59 & 2.80 & 23.75 & 31.87 \\ 
    AttMPTI~\cite{zhao2021few} & CVPR'21 
    & 12.42 & 17.93 & 3.26 & 5.52 & 25.16 & 25.50 
    & 9.24 & 12.43 & 1.68 & 2.96 & 23.82 & 31.00 \\  
    HIPO~\cite{sur2025hyperbolic} & CVPR'25 
    & 14.95 & 25.34 & 7.44 & 11.50 & 27.63 & 17.60 
    & 11.94 & 14.63 & 2.91 & 4.86 & 25.94 & 28.31 \\ 
    PIFS~\cite{cermelli2021prototype} & BMVC'21 
    & 25.39 & 34.81 & 3.43 & 6.24 & 33.11 & 8.74 
    & 24.61 & 33.66 & 3.49 & 6.32 & 33.68 & 9.88 \\ 
    CAPL~\cite{tian2022generalized} & CVPR'22 
    & 31.73 & 39.01 & 14.75 & 21.36 & 34.55 & -0.65 
    & 30.48 & 39.10 & 10.38 & 16.28 & 34.00 & -0.74 \\ 
    GW~\cite{xu2023generalized} & ICCV'23 
    & 34.27 & 41.72 & 16.88 & 23.94 & 37.67 & 1.49 
    & 33.53 & 41.85 & 14.11 & 20.99 & 37.38 & 1.36 \\ 
    \midrule
    \rowcolor{red!10} \textbf{\ourmethod~(Ours)} & -- 
    & \textbf{36.52} & \textbf{41.94} & \textbf{23.86} & \textbf{30.38} & \textbf{38.91} & \textbf{1.27} 
    & \textbf{34.78} & \textbf{41.94} & \textbf{18.09} & \textbf{25.12} & \textbf{38.46} & \textbf{1.27} \\ 
    \bottomrule
    \end{tabular}
  }
  \label{tab:scannet-full}
\end{table*}

\begin{table*}[!thb]
    \centering
    \caption{
        Comparison of baseline methods and \ourmethod on the \textit{S3DIS} dataset under the \ifsprdgm setting with $K{=}5$ and $K{=}1$.
        We report mIoU, mIoU-B, mIoU-N, HM, mIoU-I, and FPP. All metrics are higher is better except FPP. Best results are highlighted in \textbf{bold}.
    }
    \setlength{\tabcolsep}{3.5pt}
    \resizebox{\textwidth}{!}{
    \begin{tabular}{l r|rrrrrr|rrrrrr}
    \toprule
    \multirow{2}{*}{\textbf{Methods}} & \multirow{2}{*}{\textbf{Venue}} 
    & \multicolumn{6}{c|}{\textbf{$K{=}5$}} 
    & \multicolumn{6}{c}{\textbf{$K{=}1$}} \\
    \cmidrule(lr){3-8}\cmidrule(lr){9-14}
    & 
    % & \textbf{mIoU}$\uparrow$ & \textbf{mIoU-B}$\uparrow$ & \textbf{mIoU-N}$\uparrow$ & \textbf{H-mIoU}$\uparrow$ & \textbf{mIOU-I}$\uparrow$ & \textbf{FPP}$\downarrow$ & \textbf{mIoU}$\uparrow$ & \textbf{mIoU-B}$\uparrow$ & \textbf{mIoU-N}$\uparrow$ & \textbf{H-mIoU}$\uparrow$ & \textbf{mIOU-I}$\uparrow$ & \textbf{FPP}$\downarrow$ \\
    & \textbf{mIoU} & \textbf{mIoU-B} & \textbf{mIoU-N} & \textbf{HM} & \textbf{mIOU-I} & \textbf{FPP}
    & \textbf{mIoU} & \textbf{mIoU-B} & \textbf{mIoU-N} & \textbf{HM} & \textbf{mIOU-I} & \textbf{FPP} \\
    \midrule
    \rowcolor{gray!15} JT (Oracle) & -- 
    & 73.62 & 81.57 & 64.34 & 71.94 & -- & -- 
    & 73.62 & 81.57 & 64.34 & 71.94 & -- & -- \\ 
    Finetuning & -- 
    & 2.25 & 3.26 & 1.07 & 1.61 & 24.57 & 65.55 
    & 0.52 & 0.06 & 1.06 & 0.11 & 22.88 & 75.58 \\ 
    LwF~\cite{li2017learning} & TPAMI'17 
    & 12.66 & 20.29 & 3.76 & 6.35 & 37.55 & 55.36 
    & 7.89 & 12.61 & 2.39 & 4.02 & 28.44 & 63.04 \\ 
    CLIMB-3D~\cite{thengane2025climb} & BMVC'25 
    & 17.71 & 31.37 & 1.76 & 3.32 & 41.72 & 44.27 
    & 7.24 & 12.77 & 0.79 & 1.49 & 30.98 & 68.40 \\ 
    EWC~\cite{kirkpatrick2017overcoming} & PNAS'17 
    & 19.48 & 35.56 & 0.73 & 1.44 & 44.78 & 40.09 
    & 13.74 & 20.66 & 1.64 & 3.00 & 37.85 & 54.98 \\ 
    GUA~\cite{Yang_2023_CVPR} & CVPR'23 
    & 20.18 & 29.30 & 11.43 & 16.45 & 36.61 & 44.53 
    & 9.84 & 15.41 & 5.57 & 8.19 & 29.52 & 58.42 \\ 
    AttMPTI~\cite{zhao2021few} & CVPR'21 
    & 25.83 & 34.82 & 15.75 & 21.70 & 40.70 & 38.59 
    & 18.98 & 25.02 & 12.49 & 16.66 & 37.67 & 48.81 \\ 
    HIPO~\cite{sur2025hyperbolic} & CVPR'25 
    & 27.73 & 38.00 & 18.36 & 24.76 & 42.01 & 35.96 
    & 23.34 & 30.38 & 16.34 & 21.25 & 39.80 & 42.57 \\ 
    PIFS~\cite{cermelli2021prototype} & BMVC'21 
    & 40.68 & 49.96 & 29.86 & 37.38 & 53.84 & 24.93 
    & 36.13 & 50.50 & 19.36 & 27.99 & 52.19 & 24.38 \\ 
    CAPL~\cite{tian2022generalized} & CVPR'22 
    & 55.52 & 73.11 & 35.01 & 47.27 & 63.69 & 0.64 
    & 49.16 & 73.09 & 21.25 & 32.79 & 60.89 & 0.64 \\ 
    GW~\cite{xu2023generalized} & ICCV'23 
    & 57.71 & 73.38 & 39.42 & 51.29 & 63.69 & 0.04 
    & 51.73 & 73.25 & 26.62 & 39.02 & 61.15 & 0.17 \\ 
    \midrule
    \rowcolor{red!10} \textbf{\ourmethod ~(Ours)} & -- 
    & \textbf{59.41} & \textbf{73.44} & \textbf{43.03} & \textbf{54.25} & \textbf{65.24} & \textbf{-0.03} 
    & \textbf{55.36} & \textbf{73.39} & \textbf{34.32} & \textbf{46.73} & \textbf{63.02} & \textbf{0.02} \\ 
    \bottomrule
    \end{tabular}
    }
    \vspace{-0.3cm}
    \label{tab:s3dis-full}
\end{table*}

\subsection{Experimental Setup}
\label{sec:experimental_setup}
% 

% ------------------------------------------------------------
%  Datasets
% ------------------------------------------------------------

\paragraph{Datasets} 
\label{sec:dataset}
We evaluate on two standardindoor benchmarks: \textbf{S3DIS}~\cite{armeni20163d} and \textbf{ScanNet}~\cite{dai2017scannet}. 
S3DIS contains 272 scenes across six areas with 13 classes, while \textbf{ScanNet} comprises 1,513 scenes with 20 categories.
For fair comparison, following~\cite{xu2023generalized}, the six least-represented classes form the novel set~$\vecup{C}{n}$ introduced incrementally for $t{\geq}1$, while the remaining classes constitute the base set~$\vecup{C}{b}$, reflecting a long-tailed distribution. 
Additional preprocessing and incremental-stage details are provided in the \supp.

% ------------------------------------------------------------
%  Evaluation Metrics
% ------------------------------------------------------------
\paragraph{Evaluation Metrics} 
As incremental few-shot segmentation bridges generalised few-shot and incremental learning, we adopt a unified evaluation suite used in both paradigms~\cite{xu2023generalized, thengane2025climb}. 
Following the GFS protocol, we report \textit{mIoU-B}, \textit{mIoU-N}, and \textit{mIoU}, measuring segmentation performance on base, novel, and all classes, respectively. In addition, we report the harmonic mean (\textit{HM}) of \textit{mIoU-B} and \textit{mIoU-N} to assess the balance between base and novel recognition.
From the incremental perspective, we compute the average incremental mIoU (\textit{mIoU-I}), defined as the average \textit{mIoU} across all stages, and the forgetting percentage points (\textit{FPP}), measured as the drop in \textit{mIoU-B} from the base stage ($t{=}0$) to the final stage $T$.

% -----------------------------------------------------------------------------
% Implementation
% -----------------------------------------------------------------------------
\paragraph{Implementation}
For the base stage, all baselines adopt the training protocol of Xu~\etal~\cite{xu2023generalized}. During incremental stages, the backbone remains frozen, and only class-specific prototypes are updated from few-shot support samples. 
 Additional implementation details are provided in the \supp.

% ------------------------------------------------------------
%  Results
% ------------------------------------------------------------

\subsection{Results}

We evaluate \ourmethod on S3DIS and ScanNet against representative methods from multiple paradigms. Quantitative results are reported in \cref{tab:scannet-full,tab:s3dis-full}, with per-task performance in \cref{fig:tasks} and qualitative comparisons in \cref{fig:qualitative}.

\paragraph{Results on ScanNet}
\cref{tab:scannet-full} presents results on ScanNet under both $K{=}5$ and $K{=}1$ settings. \ourmethod consistently outperforms all baselines across metrics in both regimes.
Compared with the \gfsprdgm baselines CAPL and GW under $K{=}5$, our method substantially improves novel-class recognition, increasing mIoU-N from 14.75\% and 16.88\% to \textbf{23.86\%}, and HM from 21.36\% and 23.94\% to \textbf{30.86\%}. mIoU-I also improves from 37.67\% (GW) to \textbf{38.91\%}, indicating stronger stability across incremental stages.
Under the more challenging $K{=}1$ setting, \ourmethod maintains clear advantages, surpassing GW by \textbf{+3.98} and \textbf{+4.13} percentage points in mIoU-N and HM, respectively.
Relative to the recent \ifsprdgm method HIPO, our framework achieves large gains of \textbf{+16.43}, \textbf{+18.88}, and \textbf{+11.28} in mIoU-N, HM, and mIoU-I under $K{=}5$, with similarly consistent improvements of \textbf{+15.17}, \textbf{+20.26}, and \textbf{+12.52} under $K{=}1$.

\paragraph{Results on S3DIS}
\cref{tab:s3dis-full} reports results on S3DIS under $K{=}5$ and $K{=}1$. Consistent with ScanNet, \ourmethod achieves state-of-the-art performance while exhibiting minimal forgetting.
Under $K{=}5$, it surpasses the strongest \gfsprdgm baselines (GW and CAPL), improving mIoU-N from 35.01\% and 39.42\% to \textbf{43.03\%}, HM from 47.27\% and 51.29\% to \textbf{54.25\%}, and mIoU-I from 63.69\% to \textbf{65.24\%}.
In the $K{=}1$ setting, \ourmethod remains robust, achieving \textbf{34.32\%} mIoU-N and \textbf{46.73\%} HM, substantially outperforming GW (26.62\% and 39.02\%).
Compared with HIPO, our approach delivers significant gains of \textbf{+24.67} (mIoU-N), \textbf{+29.49} (HM), and \textbf{+23.23} (mIoU-I) under $K{=}5$, with further improvements of \textbf{+17.98} (mIoU-N) and \textbf{+23.22} (mIoU-I) under $K{=}1$. 

% -----------------------------------------------------------------------------
% Performance Across Tasks
% -----------------------------------------------------------------------------

\begin{figure}[!tbh]
  \centering
  \includegraphics[width=\columnwidth]{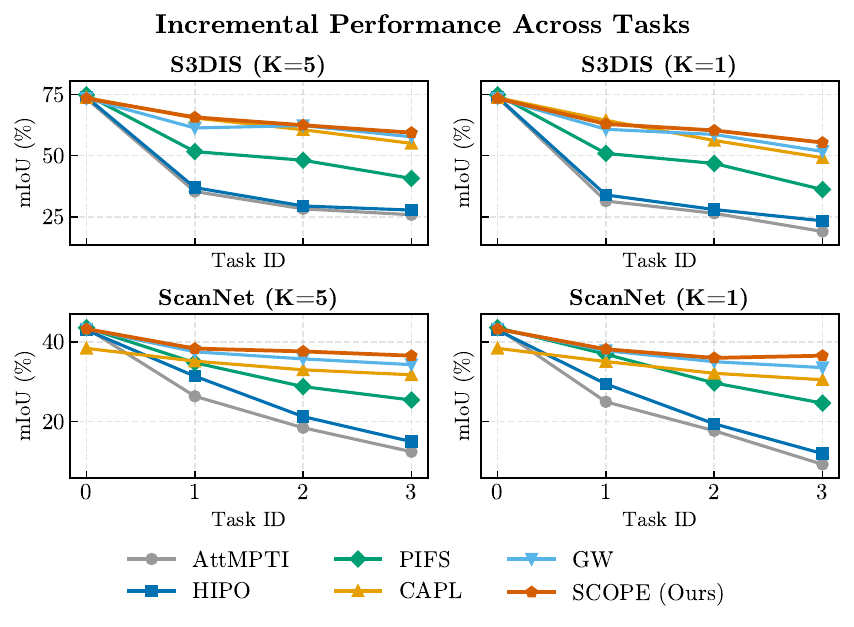}
    \caption{
      Incremental performance on S3DIS and ScanNet for $t{=}0$ to $t{=}3$ under $K{=}5$ (left) and $K{=}1$ (right), respectively. Curves show the evolution of mIoU across incremental stages.
    }
  \vspace{-0.4cm}
  \label{fig:tasks}
\end{figure}

\paragraph{Performance Across Tasks}
\cref{fig:tasks} illustrates mIoU progression over incremental tasks ($t \in \{0,1,2,3\}$) on S3DIS and ScanNet under both $K{=}5$ (left) and $K{=}1$ (right) settings. Higher trajectories at later stages indicate stronger knowledge retention and more effective adaptation to newly introduced classes.
Across all settings, \ourmethod consistently maintains a superior trajectory compared to competing approaches, with the performance gap widening as tasks accumulate--particularly under the challenging $K{=}1$ regime. This behaviour reflects a favourable stability--plasticity balance: base-class performance is preserved while novel categories are integrated with minimal degradation.
Among baselines, AttMPTI~\cite{zhao2021few} exhibits a pronounced decline across stages, consistent with its design for static few-shot learning rather than incremental updates. HIPO~\cite{sur2025hyperbolic}, despite targeting the \ifsprdgm setting, also shows noticeable performance drop-off, underscoring the difficulty of maintaining stability under limited supervision.
% 
% In contrast, the smoother and more stable progression of \ourmethod demonstrates the effectiveness of transferring scene-level contextual cues through background-guided prototype refinement, leading to reduced forgetting and more reliable incorporation of novel classes.
In contrast, \ourmethod demonstrates smoother and more stable progression across stages, evidencing the effectiveness of background-guided contextual refinement in reducing forgetting and enabling reliable incremental adaptation.

\begin{figure}[!tbh]
  \centering
  \includegraphics[width=\columnwidth]{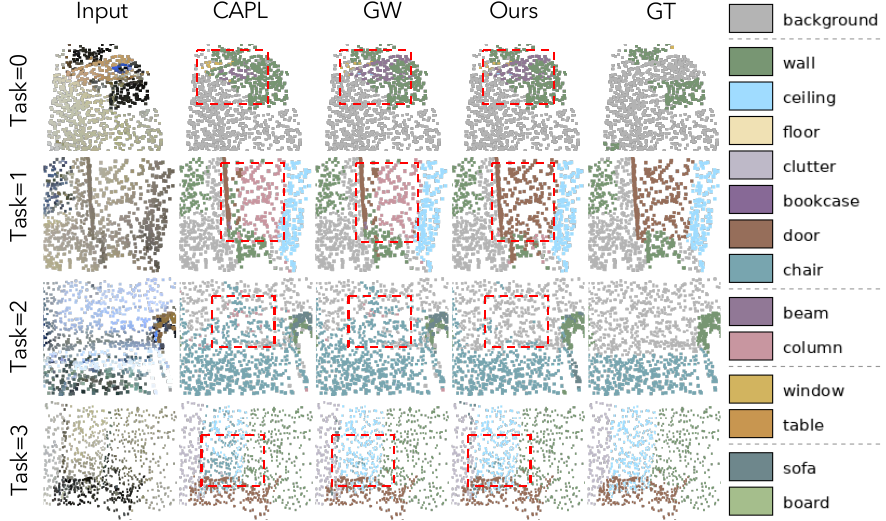}
  \caption{
    Qualitative comparison with competing methods from $t{=}0$ to $t{=}3$. The colour palette (right) denotes semantic classes, and dotted separators indicate newly introduced classes at each incremental stage (top to bottom).
  }
  \vspace{-0.4cm}
  \label{fig:qualitative}
\end{figure}

\paragraph{Qualitative Comparison}
Qualitative results from $t{=}0$ to $t{=}3$ are illustrated in \cref{fig:qualitative}, with key differences highlighted by red dashed boxes. At $t{=}0$, our method and GW produce identical predictions, as both originate from the same base model, whereas CAPL--trained under a different paradigm--exhibits visibly different outputs.
At $t{=}1$, distinctions emerge: CAPL and GW misclassify parts of the \emph{door} as \emph{column}, while our method yields more accurate segmentation with only minor errors. This trend persists at $t{=}2$, where baselines produce increasingly fragmented masks and occasionally hallucinate objects in background regions. In contrast, our predictions maintain stronger structural coherence and semantic stability.
By $t{=}3$, baseline inconsistencies become more pronounced (e.g., in the \emph{ceiling}), whereas our outputs remain better aligned with the ground truth.
Although newly introduced classes are not explicitly marked, detailed per-task visualisations and additional results--including plug-and-play demonstrations showing seamless integration with other prototype-based baselines--are provided in the \supp.
% }
% %

\noindent
Across datasets and low-shot settings, \ourmethod consistently preserves base knowledge while improving novel-class segmentation. Qualitative results show fewer artefacts and more coherent predictions across tasks, highlighting contextual cues as a key driver of incremental performance.

% ------------------------------------------------------------
%  Ablation
% ------------------------------------------------------------

\subsection{Ablation Study}
\label{sec:ablation}

\begin{table}[t]
  \centering
\caption{Comprehensive ablation study on ScanNet ($K=5$).}
  \resizebox{\linewidth}{!}{
    \begin{tabular}{lccccc}
    \toprule
    Variant & mIoU & mIoU-N & HM & mIoU-I & FPP $\downarrow$ \\
    \midrule
    Support Set Only (GW~\cite{xu2023generalized}) & 34.27 & 16.88 & 23.94 & 37.67 & 1.49 \\
    \quad + CPR (mean) & 35.68 & 22.12 & 28.91 & 38.02 & 1.50 \\
    \quad + APE (full) & \textbf{36.52} & \textbf{23.86} & \textbf{30.38} & \textbf{38.91} & \textbf{1.27} \\
    \bottomrule
    \end{tabular}
  }
  \vspace{-0.4cm}
  \label{tab:ablation_ape}
\end{table}

Below, we present ablation studies analysing the effect of CPR, APE, and sensitivity of key hyperparameters on the ScanNet dataset, following the setup described in \cref{sec:experimental_setup}.

\paragraph{Effect of CPR}
\cref{tab:ablation_ape} shows that introducing CPR on top of the GW~\cite{xu2023generalized} baseline with mean aggregation yields substantial gains on novel classes, improving mIoU-N by +5.24 and HM by +4.97 percentage points. mIoU increases from 34.27\% to 35.68\%, and mIoU-I from 37.67\% to 38.02\%, confirming the benefit of semantically aligned background prototypes for enhancing few-shot representations.

\paragraph{Effect of APE}
\label{sec:effect_of_ape}
Building upon CPR, APE further improves novel and incremental performance, increasing mIoU-N and HM by +1.74 and +1.47 points, raising mIoU-I to 38.91\%, and reducing forgetting from 1.50 to 1.27. This confirms that adaptive attention weighting suppresses noisy pseudo-instances and weak retrieval, yielding more discriminative and stable incremental learning.
% 
% ------------------------------------------------------------
%  Hyperparameters Sensitivity
% ------------------------------------------------------------

\begin{figure}[!tbh]
  \centering
  \vspace{-0.2cm}
  \includegraphics[width=\columnwidth]{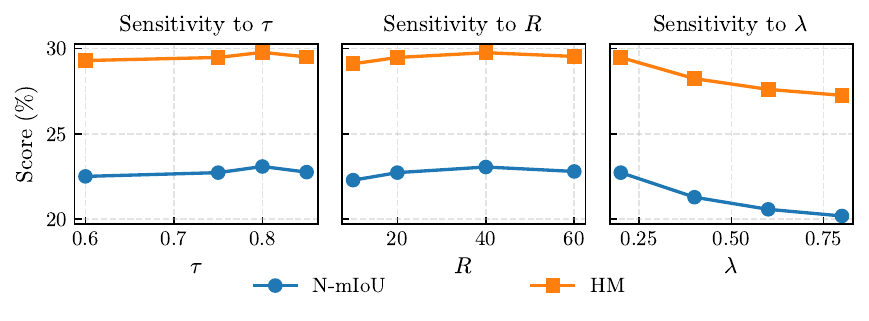}
  \vspace{-0.6cm}
  \caption{Hyperparameter sensitivity on ScanNet in terms of N-IoU and HM, obtained by varying one parameter at a time while keeping the others fixed ($\tau=0.75$, $R=20$, $\lambda=0.5$).}
  \vspace{-0.4cm}
  \label{fig:hparams_ablation}
\end{figure}

\paragraph{Effect of Hyperparameters}
We analyse three key hyperparameters: the confidence threshold $\tau$ for IPB construction, the number of retrieved prototypes $R$, and the fusion weight $\lambda$. As shown in \cref{fig:hparams_ablation}, performance remains stable across reasonable ranges. Increasing $\tau$ yields improvements up to $\tau=0.8$, after which performance slightly declines, likely due to overly aggressive filtering of informative background structures. Moderate $R$ values (around $R=40$) offer the best balance between contextual diversity and noise robustness, whereas larger $R$ may introduce less relevant prototypes. For $\lambda$, smaller values, corresponding to stronger background integration, achieve better results; as $\lambda$ increases, reduced contextual contribution leads to performance degradation, indicating that few-shot prototypes alone are insufficient for reliable \ifsprdgm.

% ------------------------------------------------------------
%  Discussion
% ------------------------------------------------------------
\subsection{Discussion}
\label{sec:discussion}

%  Error Propagation
% ------------------------------------------------------------
\paragraph{Error Propagation}
To assess the noise introduced into the IPB by the class-agnostic segmentation model, we construct an alternative IPB using ground-truth masks and compare it with the pseudo-mask version. The performance gap is small (24.77 vs. 23.86 mIoU-N and 31.20 vs. 30.38 HM for GT and pseudo masks, respectively), indicating that pseudo-mask noise has limited impact. This is largely because low-confidence instances are filtered during IPB construction, and APE further mitigates residual noise through selective prototype weighting (\cref{sec:effect_of_ape}).

% Computational Efficiency
% ------------------------------------------------------------
\paragraph{Computational Efficiency}
Compared with the strongest baseline, GW~\cite{xu2023generalized}, SCOPE incurs negligible computational overhead during incremental learning. The instance prototype bank is constructed once offline after base training, requires no further optimisation, and adds less than 1\,MB of memory storage.
During incremental stages, the only additional operations are non-parametric CPR and APE, both parameter- and training-free. As a result, the per-task computational cost remains virtually unchanged, with runtime comparable to GW (18.60\,s vs.\ 18.58\,s per task).

% Limitations [Moved to Supplementary]
% ------------------------------------------------------------
% \paragraph{Limitations}
% Although our method achieves strong performance and less sensitive to hyperparameters, it depends on the quality of pseudo-instance masks extracted during scene contextualisation. Performance may degrade if the class-agnostic model produces inaccurate or fragmented proposals. Moreover, the limited availability of class-agnostic models that do not require 3D ground-truth supervision constrains adoption, as such architectures remain scarce.

\section{Conclusion}
We introduced \ourmethod, a lightweight plug-and-play framework for \ifsprdgm. Our key insight is that background regions in base scenes encode reusable object-level structure useful for future novel classes. We exploit this by mining high-confidence pseudo-instances with a class-agnostic segmenter to build an IPB, which is used through CPR and APE to refine sparse novel-class prototypes.
This design enables adaptation under limited supervision without backbone retraining or additional parameters, achieving a strong stability--plasticity balance. Experiments on standard 3D benchmarks show consistent improvements in novel-class accuracy and reduced forgetting over representative baselines. Overall, proposed framework demonstrates that background-guided prototype enrichment is an effective and scalable strategy for low-shot continual 3D scene understanding. Future work will extend the framework to large-scale outdoor and multi-modal settings while reducing reliance on class-agnostic segmentation models.

\paragraph{Acknowledgements}
This research was supported in part by the UKRI-AHRC CoSTAR National Lab for Creative Industries Research and Development (AH/Y001060/1), the Alan Turing Mobility Grant (Academic Year 2024--26), and the Ministry of Education, Singapore, under the MOE Academic Research Fund Tier 2 (MOE-T2EP20124-0013).

{
    \small
    \bibliographystyle{ieeenat_fullname}
    \bibliography{main}
}
\clearpage
\appendix
\setcounter{section}{0}
\setcounter{figure}{0}
\setcounter{table}{0}

\renewcommand{\thefigure}{\Alph{section}\arabic{figure}}
\renewcommand{\thetable}{\Alph{section}\arabic{table}}

\setcounter{page}{1}
\maketitlesupplementary

% -----------------------------------------------------------------------------
% Overview
% -----------------------------------------------------------------------------
% 1. Competitors Comparison
% 2. Additional Dataset Details
% 3. Additional Implementation Details
% 4. Additional Experiments
% 5. Additional Discussion
% 6. SCOPE as Plug-and-Play
% 7. Per Task Breakdown

\paragraph{Overview}
This supplementary document provides additional technical details and extended analyses that complement the main paper. We first describe the datasets and preprocessing pipeline used in our experiments, including scene statistics, class distributions, and the construction of incremental tasks (\cref{sec:add_dataset}). 
We then provide additional implementation details of our framework in \cref{sec:impl_details}. 
Next, \cref{sec:plug_and_play} demonstrates the flexibility of \ourmethod by showing how the proposed Scene Contextualisation (SC) module can be integrated in a plug-and-play manner with existing prototype-based incremental learners. 
Next, \cref{sec:pertask_breakdown} reports detailed results for each incremental task, including qualitative visualisations on ScanNet and S3DIS that highlight improvements in localisation, boundary accuracy, and overall structural consistency.
Finally, \cref{sec:long_term_scalability} evaluates the robustness of \ourmethod under a long-term incremental setting spanning six stages on ScanNet, demonstrating its effectiveness in sustained incremental learning over extended task sequences. We also discuss the limitations of \ourmethod in \cref{sec:limitations}.

% ------------------------------------------------------------
% Competitor Methods Overview
% ------------------------------------------------------------
\section{Competitors Comparison}
\label{sec:competitors_comparison}
\begin{table*}[!t]
  \centering
  \caption{
    Overview of competitor methods used in our evaluation. Methods span, few-shot learning (FS), generalised few-shot learning (GFS), class-incremental learning (CI), and incremental few-shot learning (IFS) paradigms.
  }
  \label{tab:competitor_overview}
  \resizebox{\linewidth}{!}{
  \begin{tabular}{lccccl}
  \toprule
  \textbf{Method} & \textbf{Paradigm} & \textbf{Few-shot} & \textbf{Incremental} & \textbf{Prototype-based} & \textbf{Key Idea} \\
  \midrule
  FT (Na\"ive Finetuning) & -- & \xmark & \cmark & \xmark & Direct fine-tuning on new tasks without forgetting control \\
  LwF~\cite{li2017learning} & CI & \xmark & \cmark & \xmark & Knowledge distillation to preserve previous predictions \\
  EWC~\cite{kirkpatrick2017overcoming} & CI & \xmark & \cmark & \xmark & Fisher-information regularisation to protect important weights \\
  GUA~\cite{Yang_2023_CVPR} & CI & \xmark & \cmark & \xmark & Geometry and uncertainty-aware regularisation for PCS \\
  CLIMB-3D~\cite{thengane2025climb} & CI & \xmark & \cmark & \xmark & Handles long-tail distributions in incremental 3D segmentation \\
  AttMPTI~\cite{zhao2021few} & FS & \cmark & \xmark & \cmark & Multi-prototype inference with label propagation \\
  CAPL~\cite{tian2022generalized} & GFS & \cmark & \xmark & \cmark & Prototype learning with co-occurrence priors \\
  GW~\cite{xu2023generalized} & GFS & \cmark & \xmark & \cmark & Geometry-guided prototype learning for 3D segmentation \\
  PIFS~\cite{cermelli2021prototype} & IFS & \cmark & \cmark & \cmark & Prototype learning with knowledge distillation \\
  HIPO~\cite{sur2025hyperbolic} & IFS & \cmark & \cmark & \cmark & Hyperbolic prototype embeddings for incremental few-shot learning \\
  \bottomrule
  \end{tabular}}
\end{table*}

\cref{tab:competitor_overview} summarises the competitor methods considered in our evaluation. These approaches span several related paradigms, including incremental learning (specifically class-incremental learning), few-shot segmentation, generalised few-shot segmentation, and incremental few-shot segmentation. This diversity enables a comprehensive comparison under the combined challenges of data scarcity, incremental adaptation, and knowledge retention.

% ------------------------------------------------------------
% Additional Dataset Details
% ------------------------------------------------------------
\section{Additional Dataset Details}
\label{sec:add_dataset}
This section extends the dataset description provided in \cref{sec:dataset} of the main paper. We provide further details on dataset preprocessing, dataset characteristics, class partitions, and the construction of incremental learning tasks used in our experiments.

\subsection{Preprocessing}

We follow the preprocessing pipeline proposed in~\cite{zhao2021few}. Each scene is partitioned into non-overlapping $1\,\mathrm{m} \times 1\,\mathrm{m}$ blocks on the $xy$-plane, and $M{=}2048$ points are sampled from each block.  
Each point is represented by a 9-dimensional feature vector ($d_0{=}9$) consisting of the \texttt{XYZ} spatial coordinates, \texttt{RGB} colour values, and block-normalised $\overline{\texttt{XYZ}}$ coordinates.

\begin{table*}[!phtb]
    \centering
    \caption{Overview of the base and novel class partitions for S3DIS and ScanNet. 
    Following the protocol in the main paper, the majority classes form the base set $\vecup{C}{b}$,
    while the six least-represented categories constitute the novel set $\vecup{C}{n}$ introduced incrementally.}
    \vspace{0.4em}
    \begin{tabular}{p{2cm} p{8cm} p{6cm}}
        \toprule
        \textbf{Datasets} & \textbf{Base Classes} ($\vecup{C}{b}$) & \textbf{Novel Classes} ($\vecup{C}{n}$) \\
        \midrule

        \textbf{S3DIS} &
        Wall, Ceiling, Floor, Clutter, Bookcase, Door, Chair &
        Beam, Column, Window, Table, Sofa, Board \\[1.5em]

        \textbf{ScanNet} &
        Refrigerator, Desk, Curtain, Sofa, Bookshelf, Bed, Table, Otherfurniture,
        Window, Cabinet, Door, Chair, Unannotated, Floor, Wall &
        Sink, Toilet, Bathtub, Shower Curtain, Picture, Counter \\
        \bottomrule
    \end{tabular}
    \label{tab:base_novel}
\end{table*}

\subsection{Base-Novel Distribution}

\paragraph{S3DIS}
We follow the standard evaluation protocol where Area~6 is reserved for testing, while the remaining five areas provide training data for both the base and incremental stages. In addition to the overview presented in the main paper, this appendix reports the distribution of labelled points across all 13 semantic classes, summarised in \cref{tab:base_novel}. Although S3DIS is not strongly imbalanced, several categories such as \emph{beam}, \emph{column}, and \emph{board} appear less frequently than dominant structural classes like \emph{wall} and \emph{floor}. Following the protocol of~\cite{xu2023generalized}, these lower-frequency classes form the novel set~$\vecup{C}{n}$ introduced during incremental stages.

\paragraph{ScanNet}
We adopt the official split of 1{,}201 training scenes and 312 testing scenes. Compared with S3DIS, ScanNet exhibits greater variability in object occurrence across scenes, with several categories appearing only sporadically. The distribution of labelled points across all 20 semantic classes is also reported in \cref{tab:base_novel}. Following the class partitioning strategy of~\cite{xu2023generalized}, the least frequent categories are grouped into the novel set~$\vecup{C}{n}$ and introduced progressively during incremental learning.

\subsection{Incremental Task Construction}
\label{sec:incremental_splits}

In addition to the class partitions, we detail the construction of incremental tasks for both datasets. Novel classes are introduced sequentially across multiple stages following the protocol described in the main paper. Each stage introduces a small set of novel classes together with their few-shot annotated samples, simulating the gradual emergence of new object categories in dynamic indoor environments.

For clarity, we use the notation \texttt{XB-YI} to denote an incremental configuration, where \texttt{X} represents the number of base classes used during the initial training stage and \texttt{Y} denotes the number of novel classes introduced at each incremental step. For example, \texttt{15B-2I} indicates a configuration with 15 base classes and two novel classes added at every incremental stage.

\section{Additional Implementation Details}
\label{sec:impl_details}

The main experiments reported in \cref{tab:scannet-full} and \cref{tab:s3dis-full} follow a four-stage learning protocol, including the base stage. All reported results are averaged over five runs using different few-shot support samples to ensure statistical robustness.

All experiments follow the base-stage training protocol of Xu~\etal~\cite{xu2023generalized} ($t{=}0$). The encoder $\Phi$ is trained with full supervision on $\matup{D}{b}$. For incremental stages ($t{\geq}1$), the backbone remains frozen and only class-specific prototypes are updated using few-shot support sets $\matup{D}{t}$.
For the point cloud encoder $\Phi^\prime$, we adopt DGCNN~\cite{wang2019dynamic} as the backbone. The network is first pre-trained for 100 epochs and then fine-tuned during the base stage for 150 epochs. Note that our method is model-agnostic and can be readily applied to alternative backbones such as Point Transformer~\cite{zhao2021point}, as demonstrated in~\cite{xu2023generalized, Yang_2023_CVPR}.
For the scene contextualisation module, the pseudo-instance filtering threshold $\tau$ is set to $0.75$, the number of retrieved prototypes to $R{=}50$, and the fusion weight to $\lambda{=}0.5$. Although this stage is model-agnostic, we employ Segment3D~\cite{huang2024segment3d} as the class-agnostic segmenter because it is trained without 3D ground-truth supervision—a requirement that, to the best of our knowledge, is not satisfied by other currently available models.
During the novel-class registration phase, no backpropagation is performed; instead, new class prototypes are computed and integrated using the procedures described in the main paper. Other implementation details follow those of~\cite{xu2023generalized}.

\section{\ourmethod as Plug-and-Play}
\label{sec:plug_and_play}

\begin{table*}[!ht]
    \centering
    \caption{
        Effect of adding Scene Contextualisation (SC) to prototype-based \ifsprdgm methods on ScanNet (5-shot). SC consistently improves novel-class performance (mIoU-N), incremental-stage accuracy (mIOU-I), and harmonic balance (HM) across all baselines. Applying SC to GW yields our full model (\textbf{Ours}), which achieves the best overall performance.}
    \begin{tabular}{lcccccc}
    \toprule
        \textbf{Method} & \textbf{mIoU} & \textbf{mIoU-B} & \textbf{mIoU-N} & \textbf{HM} & \textbf{mIOU-I} & \textbf{FPP} \\
    \midrule
        PIFS & 25.39 & 34.81 & 3.43 & 6.24 & 33.11 & 8.74 \\
        \quad + SC & 27.36 & 34.94 & 4.93 & 8.64 & 34.03 & 8.61 \\
    \midrule
        CAPL & 31.73 & 39.01 & 14.75 & 21.36 & 34.55 & -0.65 \\
        \quad + SC & 32.97 & 39.08 & 18.70 & 25.25 & 35.19 & -0.72 \\
    \midrule
        GW & 34.27 & 41.72 & 16.88 & 23.94 & 37.67 & 1.49 \\
    \midrule
        Ours & 36.52 & 41.94 & 23.86 & 30.38 & 38.91 & 1.27 \\
    \bottomrule
    \end{tabular}
    \label{tab:play}
\end{table*}

\cref{tab:play} reports the effect of adding our Scene Contextualisation (SC) module to existing prototype-based \ifsprdgm methods on ScanNet under the $k{=}5$ shot setting. SC acts as a lightweight plug-and-play refinement that enriches prototype formation using contextual cues from background regions in base scenes, without modifying the backbone or optimisation pipeline.

Across all baselines, SC consistently improves novel-class learning while maintaining a balanced performance between base and incremental stages. For \textbf{PIFS}, SC increases mIoU-N from 3.43 to \textbf{4.93} and HM from 6.24 to \textbf{8.64}, indicating improved few-shot adaptation. \textbf{CAPL} also benefits from SC, with mIoU-N improving from 14.75 to \textbf{18.70} and HM from 21.36 to \textbf{25.25}, reflecting more stable integration of base and novel representations.
Applying SC to \textbf{GW} yields our final model, \textbf{\ourmethod}, which achieves the strongest overall performance: mIoU improves from 34.27 to \textbf{36.52}, mIoU-N from 16.88 to \textbf{23.86}, and HM from 23.94 to \textbf{30.38}, while maintaining controlled forgetting (FPP decreases from 1.49 to \textbf{1.27}). The mIoU-I also increases from 37.67 to \textbf{38.91}, indicating improved balance between knowledge retention and novel-class adaptation.

Overall, these results demonstrate that SC consistently enhances prototype-based methods, and its integration with GW produces the best-performing system. This confirms scene contextualisation as an effective and general plug-and-play component for prototype-based \ifsprdgm.

\section{Per Task Breakdown}
\label{sec:pertask_breakdown}

\begin{table*}[t]
\centering
\caption{Per-class mIoU across incremental stages on the ScanNet 15B--2I setting. 
“Mean” denotes the average over base or novel classes. Dashes indicate classes not yet introduced.}
\label{tab:incremental_breakdown}
\begin{tabular}{l c c c | c c c c c c}
\toprule
\multirow{2}{*}{\textbf{Method}} &
\multirow{2}{*}{\textbf{Task}} &
\multirow{2}{*}{\textbf{Base Mean}} &
\multirow{2}{*}{\textbf{Novel Mean}} &
\multicolumn{6}{c}{\textbf{Novel Classes} ($\vecup{C}{n}$)} \\
\cmidrule(lr){5-10}
 & & & & Sink & Toilet & Bathtub & Shower Curtain & Picture & Counter \\
\midrule

\multirow{3}{*}{GW \cite{xu2023generalized}}
& 1 & 40.92 & 13.45 & 8.96 & 17.94 & --    & --    & --    & --    \\
& 2 & 41.01 & 17.16 & 9.38 & 21.63 & 16.64 & 21.00 & --    & --    \\
& 3 & 41.72 & 16.88 & 9.78 & 22.54 & 18.16 & 22.77 & 13.78 & 14.23 \\
\midrule

\multirow{3}{*}{\textbf{\ourmethod} (Ours)}
& 1 & 41.02 & 19.29 & 12.70 & 25.88 & --    & --    & --    & --    \\
& 2 & 41.15 & 25.10 & 12.90 & 32.33 & 26.05 & 29.13 & --    & --    \\
& 3 & 41.94 & 23.86 & 17.88 & 32.59 & 25.80 & 30.13 & 17.27 & 19.51 \\
\bottomrule
\end{tabular}
\end{table*}

\begin{figure*}[!phtb]
  \centering
  \includegraphics[width=\linewidth]{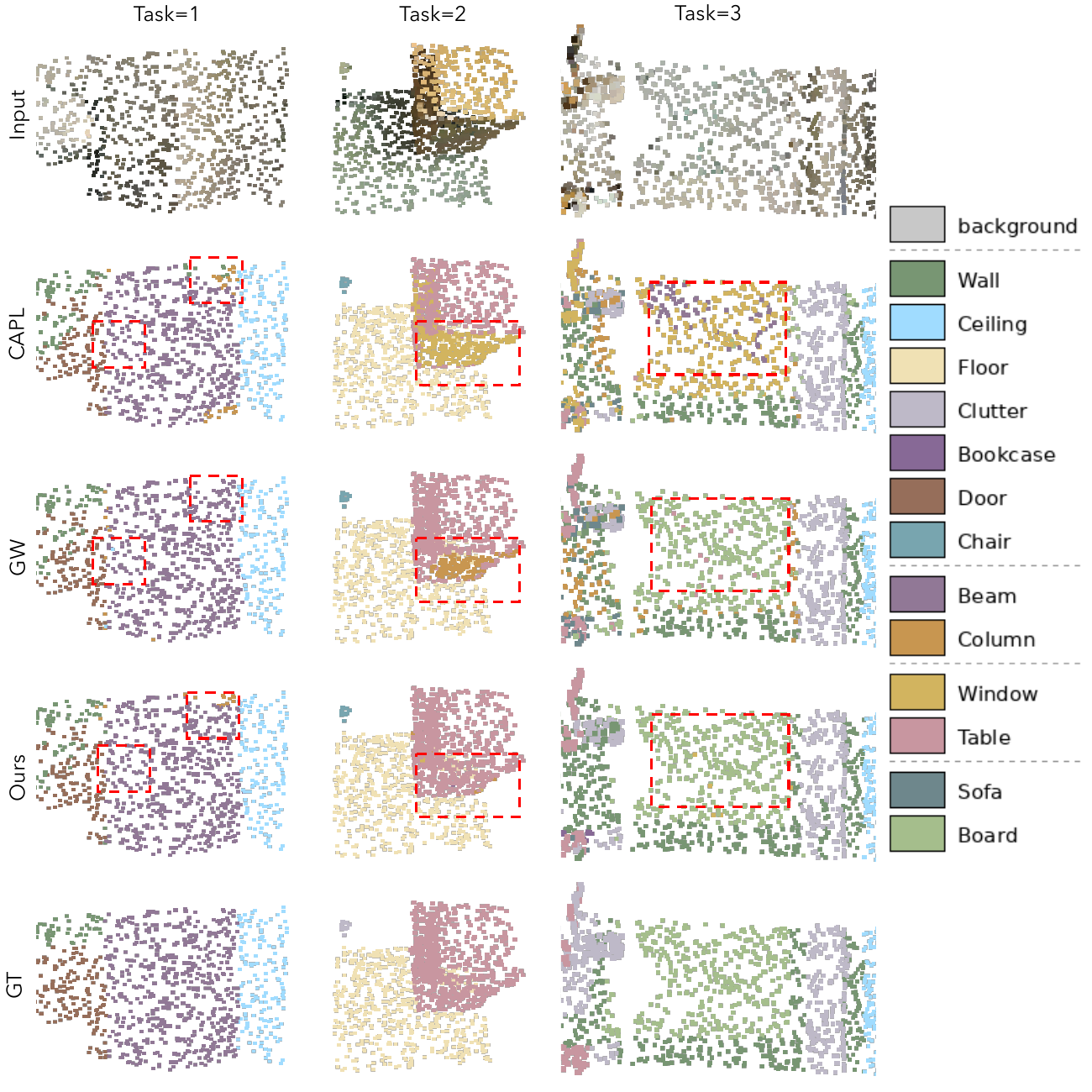}
  \caption{
    Qualitative comparison of \ourmethod with competing baselines from task $t{=}1$ to $t{=}3$. 
    The colour palette (right) denotes semantic classes, while dotted separators indicate the introduction of new classes at each incremental stage.
    % (top to bottom).
}
  \vspace{-0.4cm}
  \label{fig:s3dis_supp}
\end{figure*}

\paragraph{Incremental Performance}
\cref{tab:incremental_breakdown} presents results across all incremental stages in the ScanNet 15B--2I setting. \ourmethod consistently outperforms the previous top-performing approach, GW~\cite{xu2023generalized}. Since \ourmethod focuses on enhancing novel-class learning through prototype refinement without updating the backbone, the base-class performance remains comparable to GW across all tasks. As expected, the improvements are most pronounced on the newly introduced categories. In Task~1, \ourmethod increases the novel mean mIoU from 13.45 to 19.29, with notable gains on \emph{sink} (+3.74) and \emph{toilet} (+7.94). In Task~2, the novel mean further improves from 17.16 to 25.10, outperforming GW on all four newly introduced categories, including substantial improvements on \emph{bathtub} (+9.41) and \emph{shower curtain} (+8.13). By Task~3, after all six novel categories have appeared, \ourmethod maintains a clear advantage with a novel mean of 23.86 compared to 16.88 for GW.

Additionally, certain categories such as \emph{sink} and \emph{toilet} continue to benefit as more novel classes are introduced. For instance, the mIoU for \emph{sink} increases from 12.70 in Task~1 to 17.88 in Task~3, while \emph{toilet} improves from 25.88 to 32.59 over the same period. This trend suggests that the proposed scene-contextual prototype refinement becomes more effective as the model accumulates richer contextual cues across incremental stages.

Overall, these results show that \ourmethod not only preserves performance on base classes but also delivers significantly stronger few-shot generalisation to novel categories, enabling more robust incremental adaptation across all stages.

\paragraph{Qualitative Performance (S3DIS, 5-shot)}
As shown in \cref{fig:s3dis_supp}, under the $5$-shot setting on S3DIS, our method consistently improves as new classes are introduced, outperforming existing baselines. For example, in \textit{Task~1}, when the \textit{beam} class is introduced, CAPL frequently confuses columns with beams, while both GW and our method produce more stable predictions; however, our masks are noticeably cleaner and better aligned with the ground truth. In \textit{Task~2}, after introducing the \textit{table} class, both GW and CAPL fail to identify table regions reliably, whereas our method produces markedly more accurate and complete masks. A similar trend is observed when the \textit{board} class is added: the predictions generated by our model remain significantly closer to the ground truth, demonstrating better generalisation and reduced confusion across related categories.

\section{Long-Term Scalability}
\label{sec:long_term_scalability}
We further evaluate the robustness of \ourmethod under a long-term incremental setting spanning six stages on ScanNet. In this challenging scenario, \ourmethod achieves \textbf{19.75}/\textbf{26.79} (mIoU-N/HM), outperforming GW which attains 15.64/22.74. These results indicate that the proposed scene-contextual prototype enrichment remains effective as new classes are introduced over multiple stages, enabling stable accumulation of knowledge while maintaining competitive base-class performance. The improvements suggest that leveraging contextual cues from background regions helps the model form more reliable prototypes for novel classes, thereby supporting sustained incremental learning over extended task sequences.

\section{Limitations}
\label{sec:limitations}
Although our method achieves strong performance and is relatively insensitive to hyperparameters, its effectiveness depends on the quality of pseudo-instance masks extracted during scene contextualisation. Performance may degrade if the class-agnostic model produces inaccurate or fragmented proposals. Moreover, the limited availability of class-agnostic models that do not require 3D ground-truth supervision constrains adoption, as such architectures remain scarce.


\end{document}